\documentclass[sn-apa]{sn-jnl}

\usepackage{graphicx}%
\usepackage{multirow}%
\usepackage{amsmath,amssymb,amsfonts}%
\usepackage{amsthm}%
\usepackage{mathrsfs}%
\usepackage[title]{appendix}%
\usepackage{xcolor}%
\usepackage{textcomp}%
\usepackage{manyfoot}%
\usepackage{booktabs}%
\usepackage{algorithm}%
\usepackage{algorithmicx}%
\usepackage{algpseudocode}%
\usepackage{listings}%
\usepackage{subfig}

\theoremstyle{thmstyleone}%
\theoremstyle{thmstyletwo}%

\theoremstyle{thmstylethree}%

\begin{document}







 






\title[Crisis Talk]{Crisis talk: analysis of the public debate around the energy crisis and cost of living}


\author{\fnm{Rrubaa} \sur{Panchendrarajan}}\email{r.panchendrarajan@gold.ac.uk}

\author{\fnm{Geri} \sur{Popova}}\email{g.popova@gold.ac.uk}

\author{\fnm{Tony} \sur{Russell-Rose}}\email{t.russell-rose@gold.ac.uk}

\affil{\orgname{Goldsmiths, University of London}, \orgaddress{\city{London}, \postcode{SE14 6NW}, \country{UK}}}






\abstract{A prominent media topic in the UK in the early 2020s is the energy crisis affecting the UK and most of Europe. It brings into a single public debate issues of energy dependency and sustainability, fair distribution of economic burdens and cost of living, as well as climate change, risk, and sustainability. In this paper, we investigate the public discourse around the energy crisis and cost of living to identify how these pivotal and contradictory issues are reconciled in this debate and to identify which social actors are involved and the role they play. We analyse a document corpus retrieved from UK newspapers from January 2014 to March 2023. We apply a variety of natural language processing and data visualisation techniques to identify key topics, novel trends, critical social actors, and the role they play in the debate, along with the sentiment associated with those actors and topics. We combine automated techniques with manual discourse analysis to explore and validate the insights revealed in this study. The findings verify the utility of these techniques by providing a flexible and scalable pipeline for discourse analysis and providing critical insights for cost of living – energy crisis nexus research.}


\maketitle

\section{Introduction}

Climate change, alongside other ecological issues relating to human activity on the planet, is one of the most important, yet intractable issues facing our species at present. In no small part climate change is associated with the burning of fossil fuels, yet fossil fuels underpin much of modern prosperity and way of life. Our profound dependency on them makes climate change an extremely difficult issue to resolve, not least because it would require political and economic action on a very large scale, with potentially significant political and economic costs.

Any solutions to our fossil fuel dependency would require a public consensus around the reality of climate change, the desirability and feasibility of the required action, agreement around the nature of this action and acceptance of the cost, or at least belief that the benefit would outweigh any costs. Such public consensus has proven elusive, which is why the study of the public conversation around climate change has been the subject of numerous studies across a range of disciplines (see, for instance, \cite{boykoff11}, \cite{boykoff19}, \cite{carvalho07}, \cite{bednareketal22}, \cite{gillingsdayrell23} amongst many others). In this paper we aim to contribute to this body of research, focusing on recent media discussions of the energy crisis, here specifically in relation to the cost of living.

In the early 2020s dependency on fossil fuels has become a topic debated in a different context, that of an energy crisis in Europe, associated in part with the war in Ukraine and access to Russian oil and gas. The energy crisis could become an inflection point, potentially mobilising a turn towards a more sustainable energy policy. Such an inflection point of crisis could also become, in the words of \cite[1]{bednareketal22}, a moment of `intense discursive construction, through which individuals and communities make sense of happenings as they unfold’. The energy crisis brings into the debate geopolitical and economic interests, but, at least in the UK context, it also brings to the fore issues of social justice and equity. A study of the public debates generated around the energy crisis can help us understand the complex intersection of discourses around fossil fuels, cost of living, sustainability, and social justice.

In this paper, we report the initial results of a project that aims to deliver on these aims. We collected a sample of UK mainstream media texts (see Section \ref{dataacquisition}). The analysis is based on an up-to-date natural language processing (NLP) methodology. The next section provides the background to our methodology, and Section \ref{methods} lays out the details of our approach. Our preliminary results are described and discussed in Sections \ref{results} and \ref{discussion}. 

\section{Background} \label{background}

Understanding how the public perceives and responds to the energy crisis is of fundamental importance, and NLP offers various analytical approaches, such as topic modelling, sentiment analysis, semantic role labelling, and more. Some of these techniques operate at the level of individual tokens, while others focus on representations at the document level. In each case, it is necessary to first acquire or create a corpus of relevant documents and then to analyse that corpus.  

A good overview of corpora used by the NLP community to investigate the debate on climate change is provided in \cite{stede2021climate}. A further potentially relevant resource is the Science Daily Climate Change (SciDCC) dataset, presented in \cite{mishra2021neuralnere}, which includes approximately 11,000 news articles on the topics “Earth and Climate” and “Plant and Animals” scraped from the Science Daily website. More recently, \cite{volkanovska2023insightsnet} described the process of building the multimodal InsightsNet Climate Change Corpus (ICCC) and using NLP techniques to enrich corpus metadata, creating a dataset that supports the exploration of the interplay between the various modalities that constitute the discourse on climate change. 

Alternatively, a bespoke corpus can be created, e.g., using retrieval techniques to sample from a larger document collection or database. For example, \citealp{rebich2015image} retrieved full-text articles from the LexisNexis database using query terms identified by previous research as part of an investigation into the discourse of climate change. More recently, \cite{gillingsdayrell23} retrieved full-text articles from Factiva and LexisNexis to diachronically explore the discourse of climate change, consisting of two distinct subcorpora: a tabloid subcorpus and a broadsheet subcorpus. Similarly, \cite{liu2022climate} created two large corpora of New York Times articles by retrieving documents from LexisNexis. 
Once the corpus has been created, some degree of pre-processing is usually necessary. This will typically consist of a process of normalisation (e.g. case folding, tokenisation, stop word removal, etc.) to remove some of the linguistic ‘noise’ prior to higher-level processes such as phrase extraction and entity recognition. The normalisation process can be particularly challenging for social media data, such as tweets and other micro-blog posts, see \cite{dahal2019topic}. 

Once the data has been normalised it is possible to use a unigram bag-of-words (BOW) representation to model the individual documents (\citealp{grimmer2010bayesian}). However, this approach has the disadvantage that concepts articulated as multi-word phrases can be lost in the modelling process and are thus unavailable to downstream processes. A more robust approach is to identify and extract such phrases as part of the pre-processing and normalisation process (\citealp{handler2016bag}). There are various techniques and tools available for phrase extraction. AutoPhrase, for example, leverages high-quality phrases from public knowledge bases and utilises a POS-guided phrasal segmentation model, which incorporates the shallow syntactic information to further enhance the performance (\citealp{shang2018automated}). 

One of the more popular methods in media analysis is topic modelling, which can uncover recurring themes and subjects that shape the discourse. Topic modeling techniques, particularly Latent Dirichlet Allocation (\citealp{blei2003latent}), can be used to find patterns in many data types. In the case of climate change, this can include analysis of business sustainability reports, corporate social responsibility reports (\citealp{benites2018sustainability}), and public policy (\citealp{quinn2010analyze}). Topic modeling can help identify which entities, be they governments, organizations, or individuals, are discussed in the context of climate change responsibility. This is vital for understanding the attribution of responsibility (\citealp{jelodar2019latent}), and identifying the recurrent issues and themes within the overall discourse, the interests of various actors, and the major causes contributing to problematic issues (\citealp{benites2018topic}). Topic modeling can also be applied to track temporal changes in the prevalence of topics within climate change discourse (\citealp{blei2006dynamic}). Such approaches can reveal how discussion of the issues has evolved over time to reveal major cultural shifts, and hence provide a deeper, diachronic understanding of the problem space (\citealp{hoffman2015culture}).  

Although topic modelling is a commonly used and highly insightful technique, it is predicated on the analysis of the text at the document level. To identify the specific roles played by individual actors within the text, it is necessary first to reliably extract them and second to identify their semantic role. The first of these tasks is usually achieved by entity extraction techniques. Information Extraction (IE) is a form of text analysis which extracts structured data from unstructured text (\citealp{maynard2015understanding}). Named Entity Recognition (NER) is a key information extraction task, which is concerned with identifying instances of entities such as people, locations, and organisations. A closely related task is Named Entity Linking (NEL), which identifies repeated instances of a particular entity within a given document, or across related documents and sources (\citealp{rao2013entity}). A variety of tools exist for NER, such as StanfordNLP, NLTK, OpenNLP, SpaCy, and GATE (\citealp{schmitt2019replicable}). 

Once the entities have been extracted, it is possible to apply related techniques such as semantic role labelling to identify the particular roles played by actors and organisations within the discourse. Semantic roles are typically applied based on ‘frames’ or schemata predicated on the syntactic constructions associated with particular verbs, which in turn are a reflection of the semantic components that restrict allowable arguments (\citealp{palmer2005proposition}). For example, the verb ‘give’ would typically have three arguments: an agent (‘giver’), an object (‘thing being given’), and a beneficiary. In more sophisticated schemes, there can be different types of arguments (called `thematic roles’) such as Agent, Patient, Instrument, and also of adjuncts, such as Locative, Temporal, Manner, and Cause (\citealp{ak2018construction}). 

The above techniques assume that topics and entities can be thought of as objective concepts that are instantiated and framed dispassionately within the discourse. The reality, of course, is quite different: topics and entities within the climate change debate are the subject of much opinion and argument, often with highly polarised, contrasting sentiment. As a result, sentiment analysis techniques have also been used to explore public opinion toward climate change, categorizing opinions as positive, negative, or neutral, thus providing deeper insights into the public's emotional stance on the issue (\citealp{pak2010twitter}). Sentiment analysis can also be used in a diachronic manner, to track major changes in public sentiment, identify shifts in public perception over time, and reveal how climate change sentiment evolves (\citealp{taufek2021public}).  

The public discourse on the energy crisis and climate change is a complex narrative. NLP techniques offer a powerful lens to understand this discourse, uncovering societal attitudes, attributions of responsibility, and potential future actions. However, there are many challenges still to be overcome, and the analysis techniques usually need to be adapted to the target domain to get the best results (\citealp{derczynski2015analysis}).

\section{Methods} \label{methods}

\begin{figure}
\includegraphics[width=\textwidth]{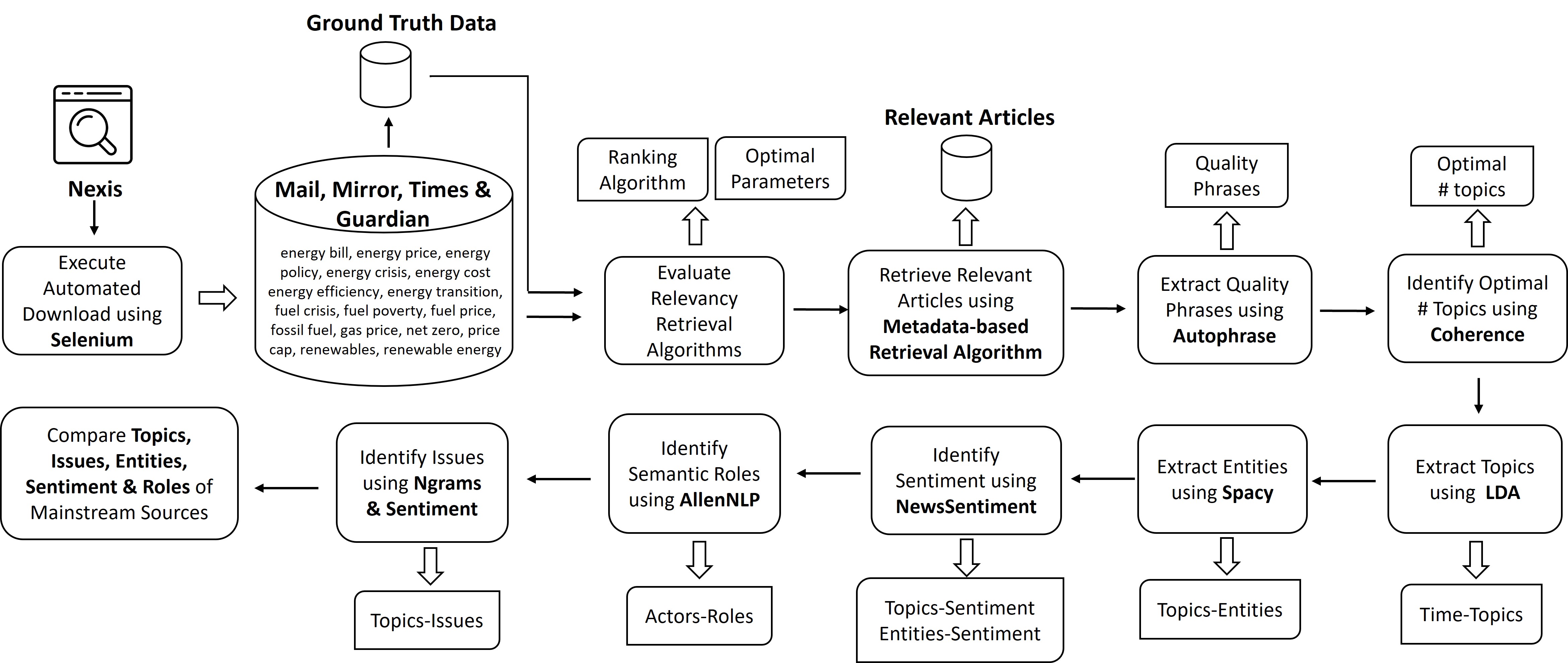}
\caption{The NLP pipeline.} \label{pipeline}
\end{figure}

To achieve wide empirical coverage and more general validity of our findings, we analysed the energy crisis discourse using automatic and semi-automatic techniques from corpus linguistics and NLP. The proposed methodology is based around a pipeline architecture composed of a set of NLP components which deliver a combination of individual insight and intermediate structure required by subsequent downstream components (see Figure \ref{pipeline}). This includes dedicated components for automated data collection, relevant article retrieval, topic modeling, entity extraction, sentiment analysis, semantic role labeling, and issue identification and visualization. These techniques have been applied to a corpus of data from mainstream media curated as part of the project, including samples from two broadsheets with different political leanings (The Times and The Guardian), and two tabloid newspapers similarly on different sides of the political spectrum (the Daily Mail and the Mirror).  The following sections explain each component of the NLP pipeline.  

\subsection{Data acquisition} \label{dataacquisition}

We collected mainstream media data from Nexis\footnote{https://www.lexisnexis.co.uk/} which offers access to a vast database of media sources. A manual analysis of articles retrieved using the search query “energy crisis” revealed that crisis talk was initiated during the period of early 2014 and gradually evolved into a central topic in the early 2020s. From the initial data collection we extracted further closely associated keywords that still referred to the main aspect of our study, i.e. the cost of energy and the impact of this cost. Thus we collected publications in The Times and The Guardian, the Daily Mail, and the Mirror during the period of 01-Jan-2014 to 31-Mar-2023 using 16 keywords (see Figure \ref{pipeline}). This resulted in a corpus of 44,168 articles.

Each article is stored in a semi-structured form by Nexis with fields for title, section, date, writer, body, and meta-data indicating the geography and subjects discussed in the body of the article. The section field generally indicates the edition of the newspaper, and we removed items from non-UK editions, i.e. Australia News, World News, and US News. To preserve the UK focus of the study, we also utilized the geography field from the meta-data. Articles mentioning any of the following terms in geography - London, England, United Kingdom, Ireland, Scotland, Wales, UK, and Ukraine were considered geographically related articles. We considered the country Ukraine as relevant geographically in order to include in the analysis the relationship between the latest crisis talk and the Ukraine war. The final corpus consists of 31,769 articles and Table \ref{stats} shows its statistics. 

\begin{table}
\caption{Corpus statistics.}\label{stats}
\begin{tabular}{|l|l|l|}
\hline
Source &  Number of Articles & Average Article Length (in Words)\\
\hline
Daily Mail & 5,089 & 678 \\
Mirror & 3,930 & 378 \\ 
The Times & 10,573 & 617 \\ 
The Guardian & 12,177 & 1,778 \\
\textbf{Total} &  31,769 & - \\ 

\hline
\end{tabular}
\end{table}

\subsection{Relevant article retrieval}
The corpus retrieved using the 16 search keywords related to energy crisis talk contained irrelevant articles as well. This is due to the retrieval nature of Nexis search, where an article with at least a single occurrence of any of the search keywords in either the title or body is retrieved as a positive hit. To sift out the irrelevant articles from the corpus, we utilized the subjects present in the metadata. Each subject listed in the metadata along with the percentage it is being discussed in the body is part of the Nexis topic taxonomy. We developed a metadata-based retrieval algorithm that starts with seed-relevant subjects and iteratively chooses relevant documents and relevant subjects until no new relevant subjects are found. The retrieval algorithm is controlled by the following three parameters. 

\begin{itemize}
    \item Discussion threshold \textit{d} – Threshold used to determine the minimum percentage of discussion of a relevant subject required to consider an article as relevant. 
    \item Popularity threshold \textit{p} – Minimum percentage of popularity of a subject among relevant articles required to consider a subject as relevant. Here the popularity is determined using the number of articles that contain the subject in meta-data. 
    \item Growth factor \textit{r} – Factor controls the growth of the discussion threshold. This enforces the increase in the percentage of discussion of a relevant subject required to consider an article as a relevant article with each iteration of the retrieval process. 
\end{itemize}

The metadata-based retrieval algorithm is presented in Algorithm \ref{algo:meta-data}. We aimed at analysing the crisis talk centered around the subjects “energy crisis” and “energy policy”, hence these keywords were used as seed relevant subjects of the algorithm.  

\begin{algorithm}
\caption{Metadata-based Relevant Article Retrieval}\label{algo:meta-data}
\hspace*{\algorithmicindent} \textbf{Input}: \textit{corpus}, \textit{d} (discussion threshold), \textit{p} (popularity threshold), \textit{r} (growth \\
\hspace*{\algorithmicindent} factor of d)\\
\hspace*{\algorithmicindent} \textbf{Output}: Relevant Articles
\begin{algorithmic}[1]
\State $relevant\_subjects \gets ["energy\:crisis", "energy\:policy"]$
\State $new\_subjects \gets relevant\_sub$
\State $relevant\_docs \gets [] $
\State $t \gets 0 $
\While{$new\_subjects$ is not empty}
    \For{each $article$ in $corpus$}
    \If{$article$ contains any $new\_subjects$ with discussion \% $ \geq d*r^t$ }
    \State Add $article$ to $relevant\_docs$
    \EndIf
    \EndFor
    \State Remove $relevant\_docs$ from $corpus$ 
    \State $new\_subjects \gets [] $
    \For{each $subject$ in $relevant\_docs$}
    \If{$subject$ not in $relevant\_subjects$ and popularity in $relevant\_docs$ $\geq p$ }
    \State Add $subject$ to $new\_subjects$
    \EndIf
    \EndFor
    \State Add $new\_subjects$ to $relevant\_subjects$
    \State $t \gets t + 1 $
\EndWhile
\end{algorithmic}
\end{algorithm}

We created a human-annotated ‘gold standard’ validation set for choosing the optimal parameters of the metadata-based document algorithm and comparing it with traditional document ranking algorithms. We randomly sampled 150 articles from the corpus and labeled them as “relevant” or “irrelevant” with respect to the subjects “energy crisis” and “energy policy”. Table \ref{validationsetstats} presents the statistics of the validation set. It can be observed that only 61.3\% of the sample is relevant to the subjects “energy crisis” and “energy policy”, thus showing the need to retrieve the relevant articles from the whole corpus for accurate results from the remaining pipeline. 

\begin{table}
\caption{Statistics of validation set.}\label{validationsetstats}
\begin{tabular}{|l|l|l|l|}
\hline
Source &  $\#$ relevant docs & $\#$ irrelevant docs & Total\\
\hline
Daily Mail & 27 & 18 & 45 \\
Mirror & 28 & 10 & 38 \\ 
The Times & 24 & 15 & 39 \\ 
The Guardian & 13 & 15 & 28 \\ 
Total   & 92  (61.3\%) &   58 &  150 \\ 
\hline
\end{tabular}
\end{table}

\begin{table}
\caption{Performance of the Relevant Article Retrieval Algorithms.}\label{performance}
\begin{tabular}{|l|l|l|}
\hline
 &  F1 Score & Number of relevant documents \\
\hline

Metadata-based Retrieval (d=70, p=25, r=1) &
0.657 & 
16,589 (52.2\%) \\
Word embedding-based Retrieval &
0.554 &
28,729 (90\%) \\ 
TF-IDF-based Retrieval & 
0.625 & 
20,868 (65.7\%) \\ 

\hline
\end{tabular}
\end{table}

We varied the parameters of the metadata-based retrieval algorithm as follows: discussion threshold and popularity threshold varied from 0 to 100 in a step size of 5, and the growth factor varied from 1 to 1.4 in a step size of 0.1. This resulted in 2100 combinations of parameters. We compared the meta-data-based article retrieval algorithm with the traditional TF-IDF and Word-embedding-based ranking techniques\footnote{https://github.com/4OH4/doc-similarity}. We observed that the metadata-based retrieval algorithm outperformed TF-IDF-based and Word-embedding-based retrievals. F1 scores of the three algorithms and the optimal setting of the metadata-based retrieval algorithm are listed in Table \ref{performance}. Table \ref{relsubj} presents the relevant subjects in the order they were retrieved and the number of relevant articles from each newspaper source. Subjects related to energy shortage, fuel price, and inflation are commonly retrieved as relevant subjects across all four sources. All the common relevant subjects are highlighted in Table \ref{relsubj}. 

\begin{table}[]
    \caption{Relevant Subjects from the Relevant Articles Retrieved.}\label{relsubj}
    \begin{tabular}{|p{2cm}|p{10cm}|}
    \hline
    Source & Relevant Subjects \\
    \hline
         Daily Mail (3198, 62.8\%) & \textbf{energy shortages, price increases, energy \& utility policy, oil \& gas prices, prices, inflation, cost of living}, taxes \& taxation \\
    \hline
    Mirror (2165,55\%) & \textbf{energy shortages, energy \& utility policy, price increases, oil \& gas prices, prices, inflation}, cost of living  \\
    \hline
    The Times (5477,51.8\%) & \textbf{energy shortages, energy \& utility policy, oil \& gas prices}, public policy, energy \& utility regulation \& policy, \textbf{prices, inflation, price increase}  \\
    \hline
    The Guardian (5749, 47.2\%) & \textbf{energy shortages, energy \& utility policy, oil \& gas prices}, public policy, energy \& utility regulation \& policy, \textbf{prices, price increases, inflation}  \\
    \hline
    \end{tabular}
\end{table}

\subsection{Phrase Mining}
This component receives the relevant articles retrieved using the metadata-based article retrieval algorithm as input and generates quality phrases to produce an intermediate structure for the downstream components. Considering only the individual words or all the possible n-grams of the corpus may not lead to an effective and scalable information retrieval pipeline. To serve this purpose, we used Autophrase (\citealp{shang2018automated}), a phrase mining tool for extracting quality phrases from very large corpora. Quality phrases are extracted by Autophrase based on four factors: popularity, concordance, informativeness, and completeness. Given an input corpus, the tool generates quality scores for phrases ranging from single words to 6-grams with respect to the surrounding words and annotates phrases exceeding the thresholds as quality phrases. The default implementation of the tool uses 0.5 and 0.8 as quality thresholds for individual words and phrases respectively.  

\subsection{Topic Extraction}
Identifying key issues discussed in the corpus is an essential task to analyze the discussions centered on these issues. In pursuit of this goal, we used a topic modeling approach to identify topics discussed in the corpus. We initially considered the extraction of hierarchical topics based on the hypothesis that the discussion in the corpus can be structured into root, super topics, and sub-topics each addressing various aspects of the energy crisis with varying degrees of depth. However, the initial results were not promising, revealing the non-hierarchical nature of topics discussed in the corpus. Following that, we used Latent Dirichlet Allocation (\citealp{blei2003latent}) to extract flat topics from the corpus. 

\begin{figure}[!ht]
        \subfloat[the Daily Mail]{%
            \includegraphics[width=.5\linewidth]{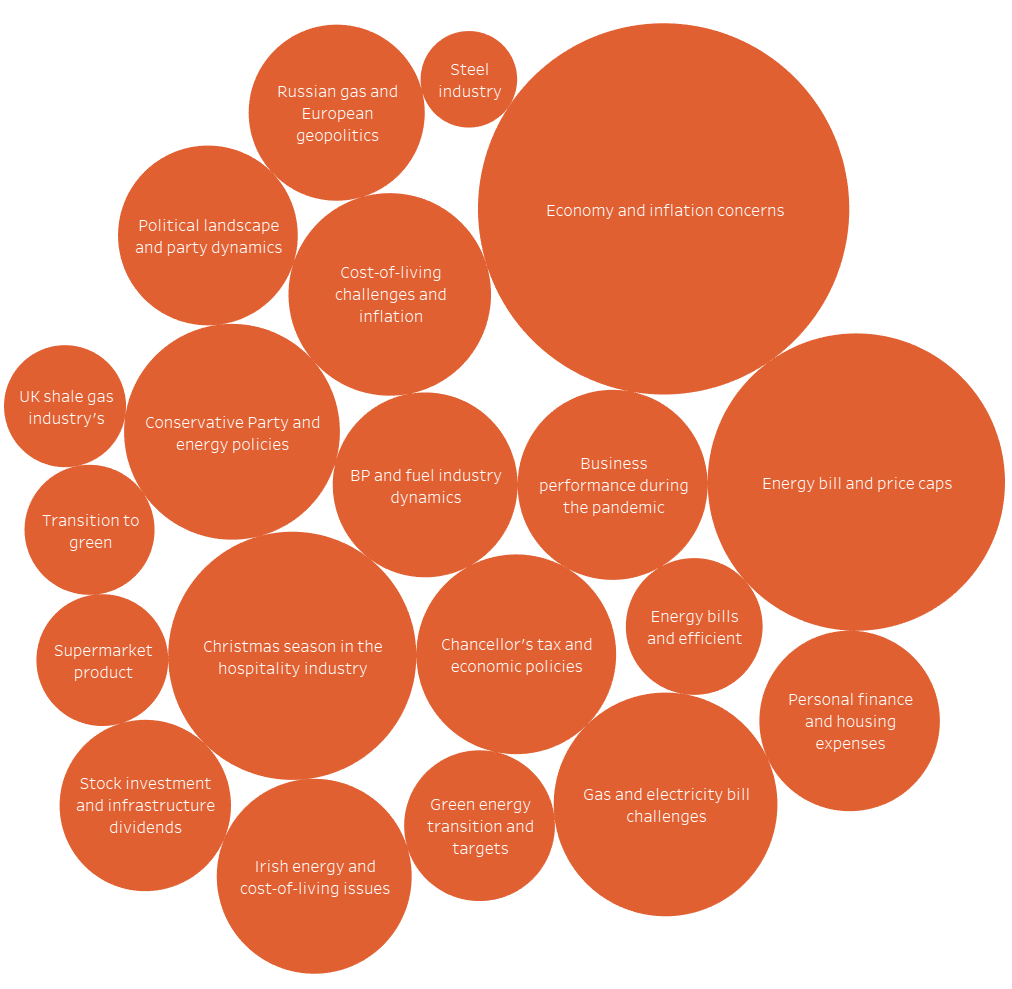}%
            \label{subfig:a}%
        }\hfill
        \subfloat[the Mirror]{%
            \includegraphics[width=.5\linewidth]{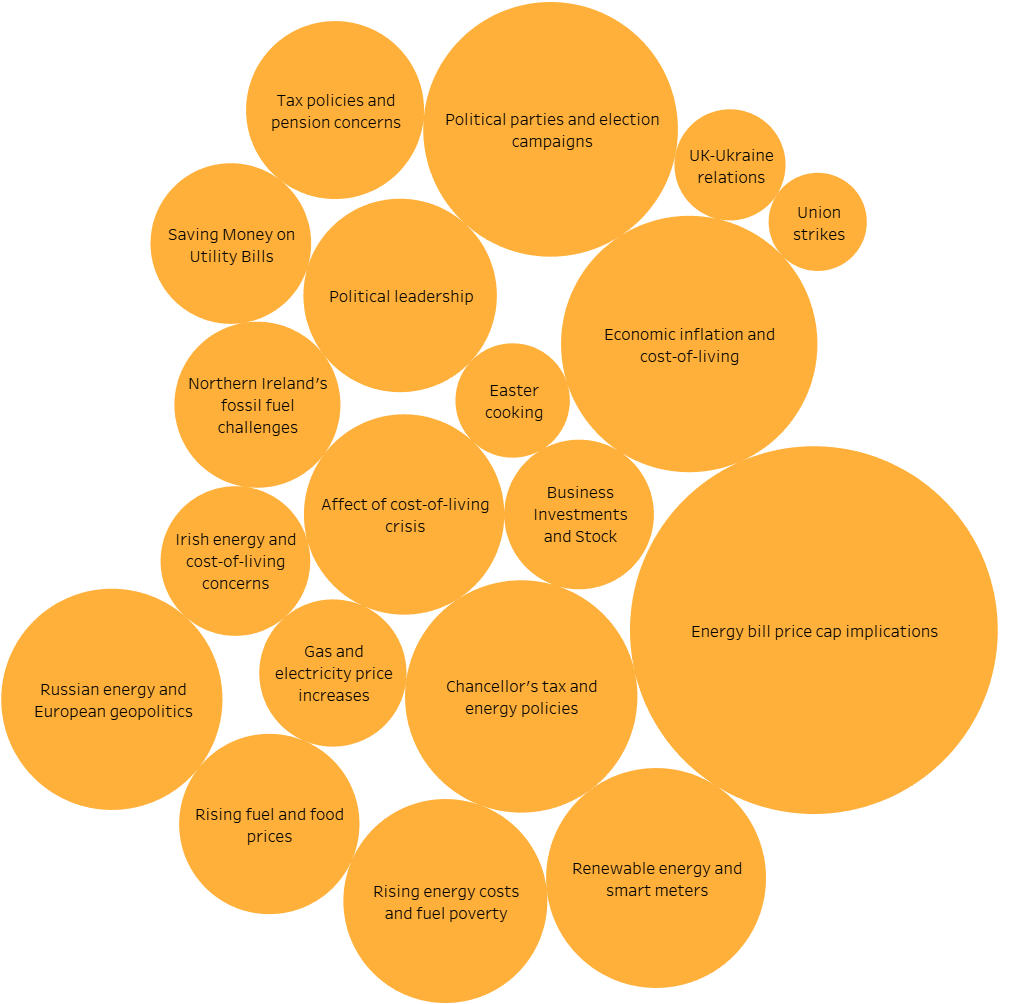}%
            \label{subfig:b}%
        }\\
        \subfloat[The Times]{%
            \includegraphics[width=.5\linewidth]{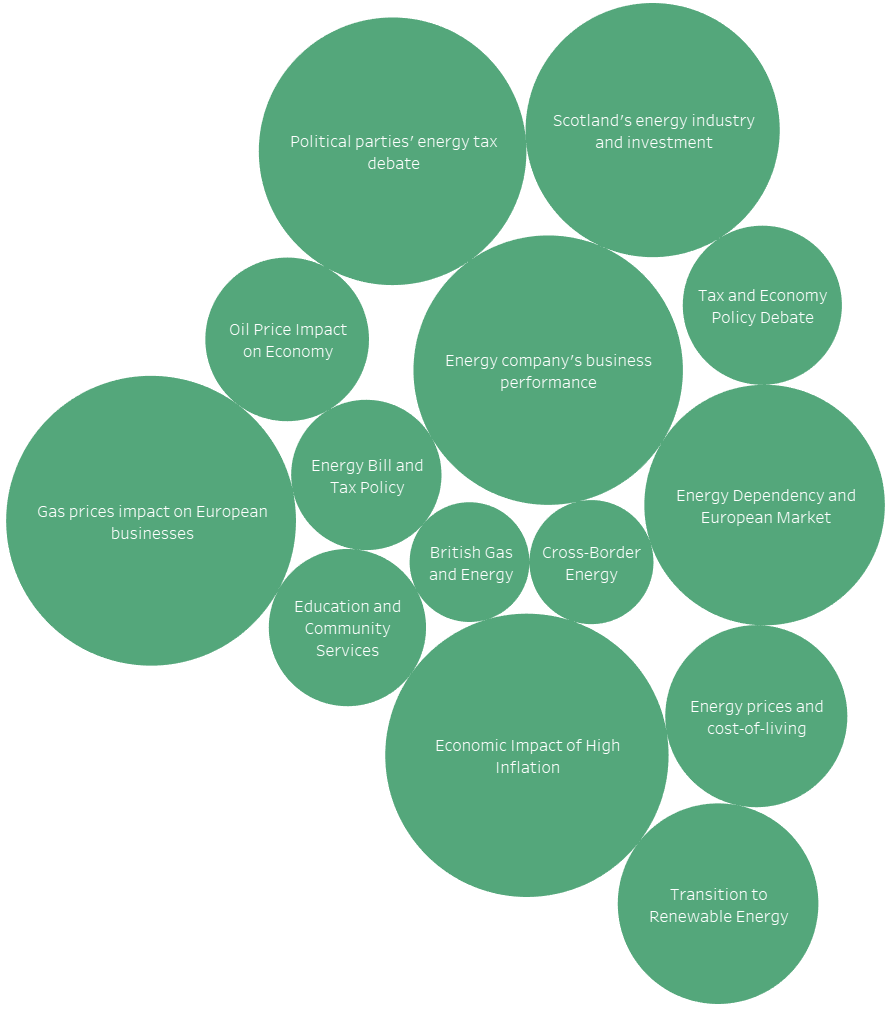}%
            \label{subfig:c}%
        }\hfill
        \subfloat[The Guardian]{%
            \includegraphics[width=.5\linewidth]{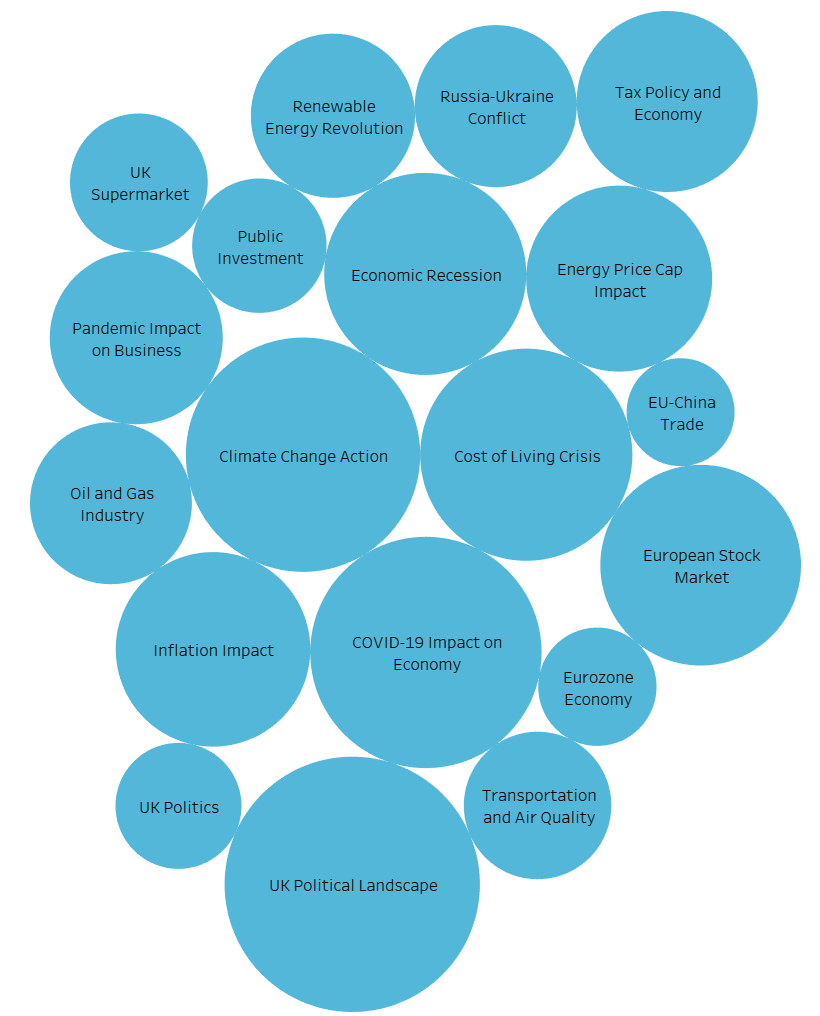}%
            \label{subfig:d}%
        }
        \caption{Topics learned across four sources}
        \label{fig:topic-labels}
\end{figure}

We used coherence (\citealp{roder2015exploring}) as an evaluation metric to determine the optimal number of topics for each source.  Coherence measures the degree of semantic similarity between the top N words of a topic. While there are different versions of coherence metrics available, we use the CV measure which is shown to have a high correlation with human judgment (\citealp{roder2015exploring}). Following previous studies (\citealp{benites2018topic}), we range the number of topics from 10 to 20 and obtain the coherence score as an average of three runs, each composed of 500 epochs. Table \ref{table:optimal-topics} shows the optimal number of topics identified for each source. Once the topics were learned, we used chat-GPT (\citealp{openai2023gpt4}) to generate a topic label for each topic using the top 20 words of the topic. Chat-GPT was prompted with the top 20 words to identify a topic label with a maximum of 5 words. Figure \ref{fig:topic-labels} presents the topic labels of the topics learned (we discuss this further in Section \ref{results}).   

\begin{table}
\caption{Number of Optimal Topics.}\label{table:optimal-topics}
\begin{tabular}{|l|l|}
\hline
Source &  Number of Topics \\
\hline
Daily Mail & 20 \\
Mirror & 19 \\ 
The Times & 14 \\ 
The Guardian & 20 \\ 
\hline
\end{tabular}
\end{table}

\subsection{Entities, Sentiment and Role Extraction}
Another key objective of this analysis is to identify key actors and their roles with respect to the issues in crisis talk. We used Spacy\footnote{https://spacy.io/} for identifying the actors, by extracting the named entities, and sentiment expressed towards the actors was extracted using NewsSentiment (\citealp{hamborg2021newsmtsc}). NewsSentiment performs target-dependent sentiment analysis, and we input actors as targets to the sentiment model to identify the sentiment expressed towards an actor as a score ranging from –1 to 1. Actors and sentiments in a sentence are linked to a topic by assigning a topic label at the sentence level, based on major topics discussed in a sentence. The average sentiment score of an actor or a topic is obtained by taking the mean value of all the sentiment scores associated with it.   

To identify the roles played by the actors, we performed semantic role labeling using AllenNLP (\citealp{Gardner2017AllenNLP}). Among the arguments extracted by AllenNLP, we use the agent or causer of a verb (indicated as ARG0) and the target (indicated as ARG1) for role analysis of actors. For example, Table \ref{table:agent-example} and Table \ref{table:target-example} present the top 5 agents and targets of the topic “Energy bill and price caps” from the Daily Mail and the top 10 verbs they are involved in the corresponding roles. 

\begin{table}[]
\caption{Top 5 Agents and their corresponding popular verbs of the topic “Energy bill and price caps” from the Daily Mail}\label{table:agent-example}
\begin{tabular}{|l|p{10.3cm}|}
\hline
Agent      & Top 10 verbs                                                                       \\ \hline
Government & needs, take, act, cover, introduce, announced, set, pay, put, steps                \\ \hline
households & struggling, save, paying, face, pay, switching, facing, reduce, see, paid          \\ \hline
customers  & pay, save, switch, switching, make, struggling, get, use, keep, cut                \\ \hline
people     & pay, save, need, switch, struggling, cut, make, switching, feel, thought           \\ \hline
suppliers  & charge, offer, charging, pass, offering, cut, ensure, hitting, calculated, sending \\ \hline
\end{tabular}
\end{table}

\begin{table}[]
\caption{Top 5 Targets and their corresponding popular verbs of the topic “Energy bill and price caps” from the Daily Mail}\label{table:target-example}
\begin{tabular}{|l|p{10.2cm}|}
\hline
Target       & \multicolumn{1}{c|}{Top 10 verbs}                                                         \\ \hline
bills        & rise, rising, soaring, cut, soar, reduce, increase, pay, fall, estimated                  \\ \hline
customers    & existing, moved, protect, transferred, leave, urged, panic, transfer, misleading, worried \\ \hline
prices       & rise, fall, rising, raising, pushed, raised, increased, pushing, raise, cut               \\ \hline
energy       & used, save, buy, use, switching, saving, provide, conserve, costs, imported               \\ \hline
energy bills & rise, rocketing, soaring, freeze, soar, cap, rising, jumped, cut, reduce                  \\ \hline
\end{tabular}
\end{table}

\subsection{Issue Identification}
We define an issue as a popular noun or verb phrase with extreme variation in sentiment (\citealp{choi2010identifying}). Accordingly, we considered topic labels as default candidate issue phrases representing root issues in the corpus. We further extracted n-grams from the top 100 words of each topic as candidate phrases for issue identification. We computed the average positive, and average negative sentiment score of each candidate phrase and classified them as an issue if the difference between average positive sentiment and average negative sentiment is greater than 0.8. As anticipated, all the topics were classified as issues due to the highly polarized nature of the topic discussions. In addition to that, the issues identified using popular n-grams unveiled both unique issues associated with a single topic as well as the common issues, being the central discussion across multiple topics. For example, Table \ref{table:issue-example} shows sample topics and issues identified in the Daily Mail.  While `global warming' is an issue specific to the topic `Transition to green transportation solutions', `cost of living' can be observed as a common issue across multiple topics. 

\begin{table}[]
\caption{Sample topics and issues identified in the Daily Mail}\label{table:issue-example}
\begin{tabular}{|p{3.5cm}|p{8.5cm}|}
\hline
Topic  & Sample Issues                                                                   \\ \hline
Economy and inflation concerns               & base rate, borrowing costs, cost of living, economic growth, financial crisis, living standards, price rises, raise rates, rising prices, soaring inflation \\ \hline
Energy bill and price caps                   & annual bills, customer service, direct debit, energy price cap, fuel poverty, loyal customers, price cap, price hikes                                       \\ \hline
Personal finance and housing expenses        & base rate, cost of living, council tax, higher rate, home loan, household bills, savings rates                                                              \\ \hline
Transition to green transportation solutions & carbon emissions, charging points, electric car, global warming, net zero, petrol and diesel                                                                \\ \hline
Chancellor's tax and economic policies       & basic rate of income tax, economic growth, fiscal policy, higher taxes, income tax, living standards, small businesses                                      \\ \hline
\end{tabular}
\end{table}

\subsection{Visualization}
The final component of the pipeline aggregates the information retrieved to generate conceptual graphs that we can visualize to illustrate and explore the relationships between topics, issues, entities, and sentiment. We consider actors extracted using Named entity recognition and agents and targets extracted using semantic role labeling as entities of the system. We used Pyvis\footnote{https://pyvis.readthedocs.io/}, a Python library to generate the following four categories of conceptual graph: 
\begin{enumerate}
    \item Topics-Issues Graph – Depicts the relationship between topics and issues of a particular source and the properties of an issue including overall and topic-level popularity. 
    \item Topics-Entities Graph – Depicts the relationship between topics and entities of a given source and the properties of an entity including popularity, role, and sentiment. 
    \item Source-Issues Graph – Depicts the top 50 issues across all sources and the properties of an issue including popularity and sentiment. 
    \item Issue-Entities Graph – Depicts the popular entities across all sources for a given issue and the properties of those entities including popularity, role, and sentiment. 
\end{enumerate}

The size of a node is used to roughly indicate its overall popularity and the size of an edge is used to roughly indicate the popularity of the node with respect to the connected node (e.g., edge size between an issue node and topic node roughly indicates the popularity of an issue within the topic discussion). The Color of an edge is used to indicate the sentiment expressed towards the node with respect to the connected node (e.g., the color of an edge between an entity node and topic node indicates the average sentiment expressed towards an actor within the topic discussion). The following color codes are used to represent sentiment via edges in the concept graphs: 

\begin{itemize}
    \item Red: Negative sentiment 
    \item Green: Positive sentiment 
    \item Blue: Neutral sentiment 
    \item Gray: No sentiment extracted 
\end{itemize}

Further, we use color codes for the entity nodes to indicate the role of the entities in concept graphs as follows.  
\begin{itemize}
    \item Black: Entity plays the role of the agent only 
    \item Yellow: Entity plays the role of the target only 
    \item Purple: Entity plays the roles of both agent \& target 
    \item Light Blue: Entity plays no role 
\end{itemize}

\section{Results} \label{results}

By identifying topics, the NLP analysis allows us to gain an understanding of where the media directed its attention during the energy crisis. As the topics were given labels separately for each source and sometimes different topics have similar labels (e.g. `Eurozone economy' and `European stock market' in The Guardian), it makes sense first to amalgamate some topics manually into broader themes and then try to compare across the different newspapers.  
 
Topics across all newspapers fall into five broad themes: (i) energy and cost of living, (ii) economy, (iii) politics, (iv) geopolitics, and (v) climate change and green energy. That the main framing of the energy crisis is as a cost-of-living issue is not surprising -- energy cost was one of the main contributors to substantial inflationary pressures experienced in the UK during this period of time (see ONS data) -- though as noted later, different newspapers give different prominence to this aspect. The topics of economy, politics, and geopolitics are interconnected, with some topics falling more clearly into one of these themes, and some combining all of them. The energy crisis in the UK was one contributor to an economically turbulent time given the economic effects of the COVID-19 pandemic (a topic in The Guardian), slowing growth (`recession' is a topic in The Guardian, `economic impact of high inflation' is a topic in The Times), and political events during, for instance, the cabinet of Liz Truss (`tax and economy policy debate' is a topic in The Times). The theme of the economy covers some cross-border issues as well, with the `European stock market' appearing as a topic in The Guardian, for instance. The Guardian more generally seems to pick up more European issues in comparison to other newspapers, and this interest in Europe is partially confirmed by the announcement recently of an Autumn 2023 launch of a European issue of The Guardian. There also seems to be more discussion in The Guardian of the impact on the economy of the pandemic. The Guardian and the Mail also pick up the theme of the food industry. This does not seem to be as prominent in the Mirror and The Times, so is not picked up as a separate topic. 
  
As we had anticipated, climate change is another strand of the public debate. It seems to be covered with more intensity in The Guardian, where it is a separate topic, but all newspapers pay attention to the related topics of renewable and green energy. There are some suggestions from previous research that at times of crisis the topic of climate change tends to be moved to the background and the intensity of discussion in the media reduces (\citealp{boykoff11}). Topic modelling here suggests that if the energy crisis had that effect in this case, it was less pronounced for The Guardian. But it also suggests that the general focus is on replacing current energy sources with more climate-friendly ones. Energy itself is a prominent topic across all newspapers, with topics suggesting some attention paid to the oil and gas industry, to the impact of the energy crisis (i.e. to the general issue of a cost-of-living crisis and inflationary pressures), and to the various measures taken to deal with it (e.g. energy price cap) and their impact. There are also some variations here, however. For instance, the Mail seems to focus more attention on the UK shale gas industry, the steel industry, and nuclear power, which come up as separate topics. 
  
Perhaps the most visible difference identified in the topic modelling part of the analysis is how much less variation there is in the Mirror. There are fewer topics, and the ones identified seem to relate predominantly to the cost-of-living crisis. Where other newspapers discuss the energy crisis as a political, economic, and geopolitical problem, the main, if not exclusive, framing in the Mirror, it would seem, of the energy crisis is as a cost-of-living issue. 
  
\begin{figure}
\includegraphics[width=\textwidth]{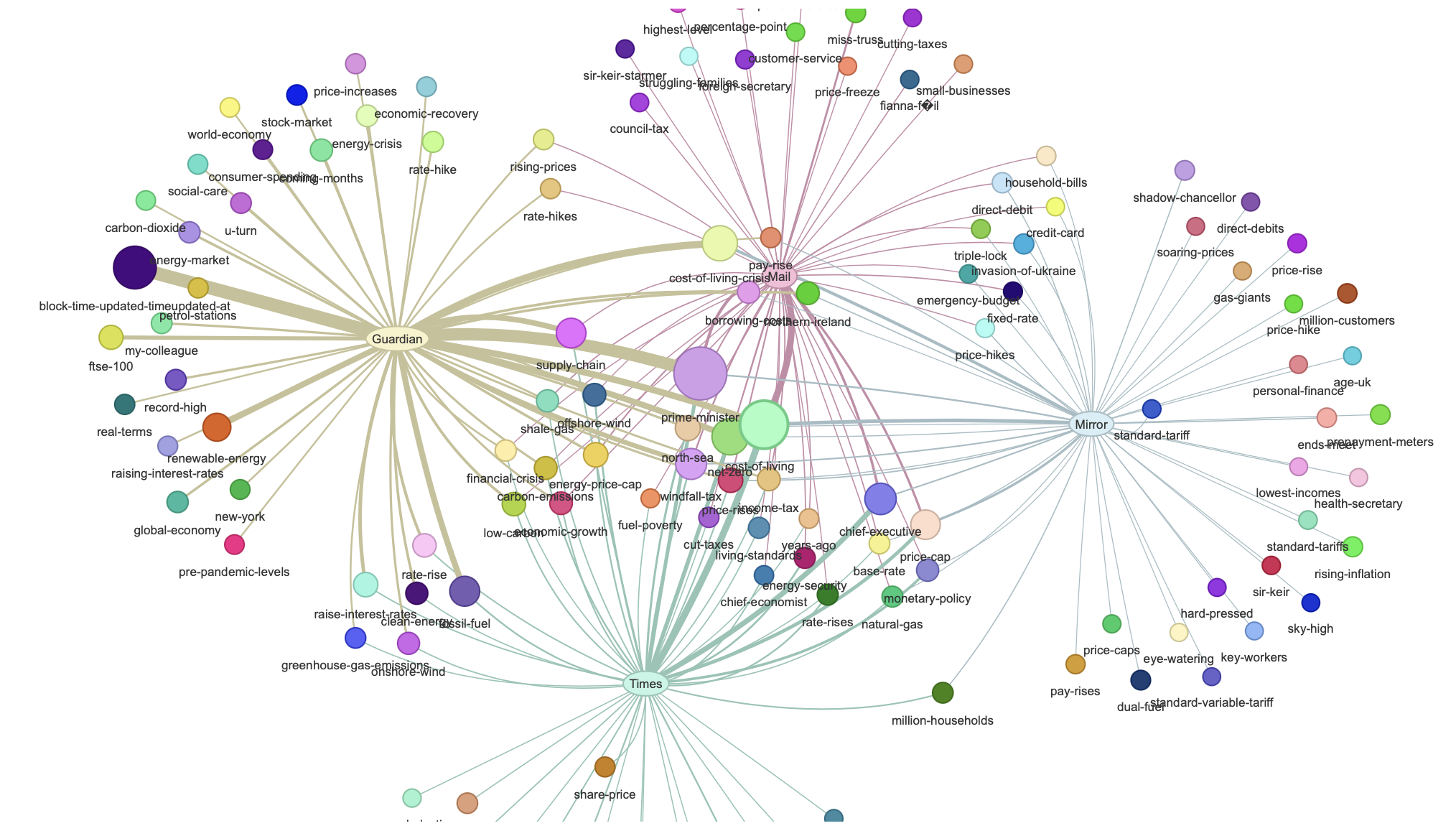}
\caption{Issues across all four sources.} \label{issues4sources}
\end{figure}

\begin{figure}
\includegraphics[width=\textwidth]{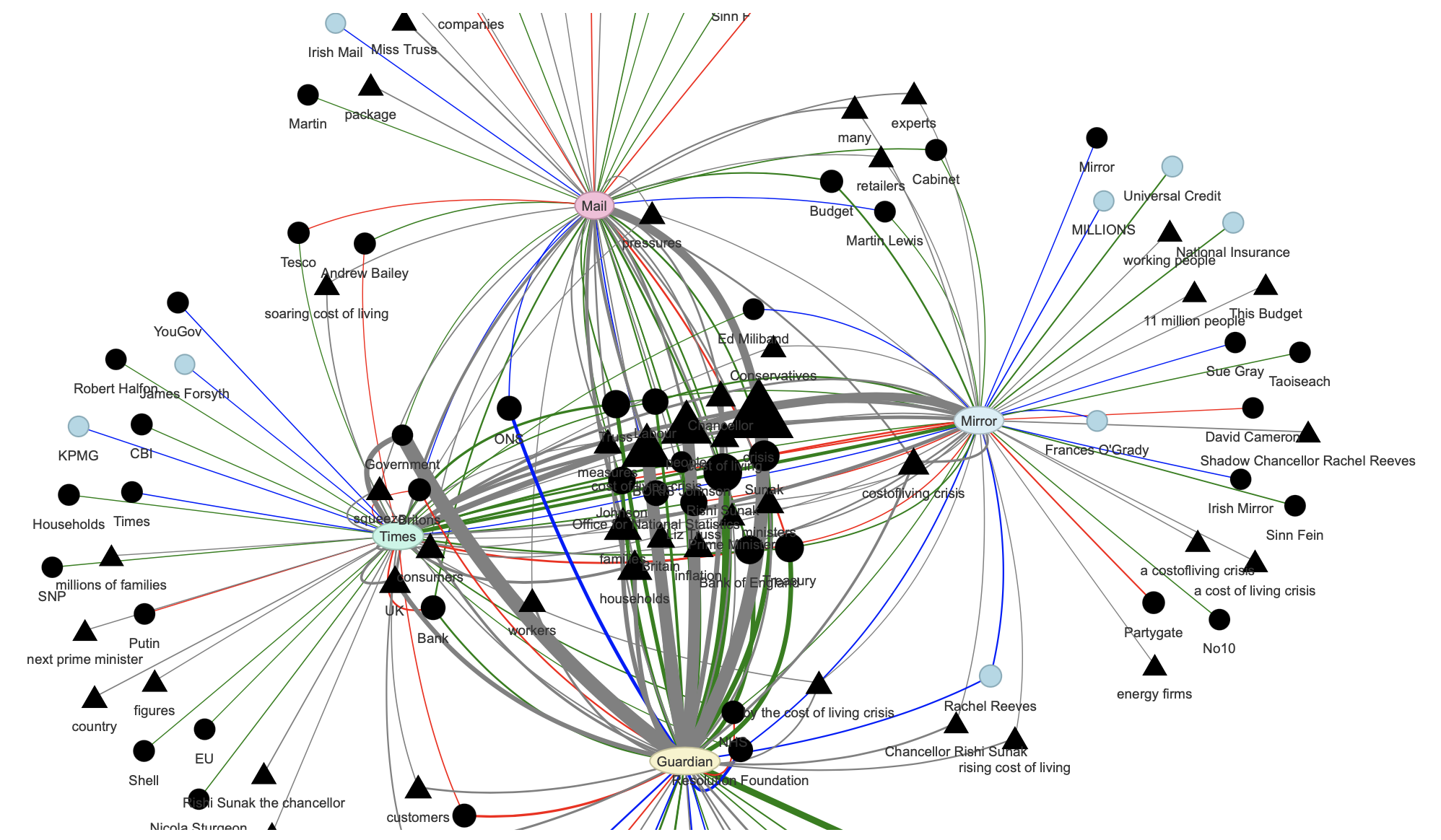}
\caption{The `cost-of-living' issue and actors associated with it.} \label{costofliving}
\end{figure}

Some of these observations can be elaborated by looking at particular issues, as defined in Section \ref{methods}. We can trace which issues are shared by which newspapers, and which are not, see Figure \ref{issues4sources}. 
Other visualisations allow us to zoom in on particular issues and the actors associated with them. For instance, one issue all newspapers give attention to is the cost-of-living, see Figure \ref{costofliving}. 
As is clear from Figure \ref{costofliving}, the cost of living is framed as a political/policy problem, with the actors being either those affected by it (households, people, families), and those called upon to provide solutions (Conservatives, Bank of England, Johnson). On some of these actors the media seems politically divided, e.g. in this graph Liz Truss and Sunak attract negative sentiment from the Mirror, but positive sentiment from the other three newspapers.  
 
\begin{figure}
\includegraphics[width=\textwidth]{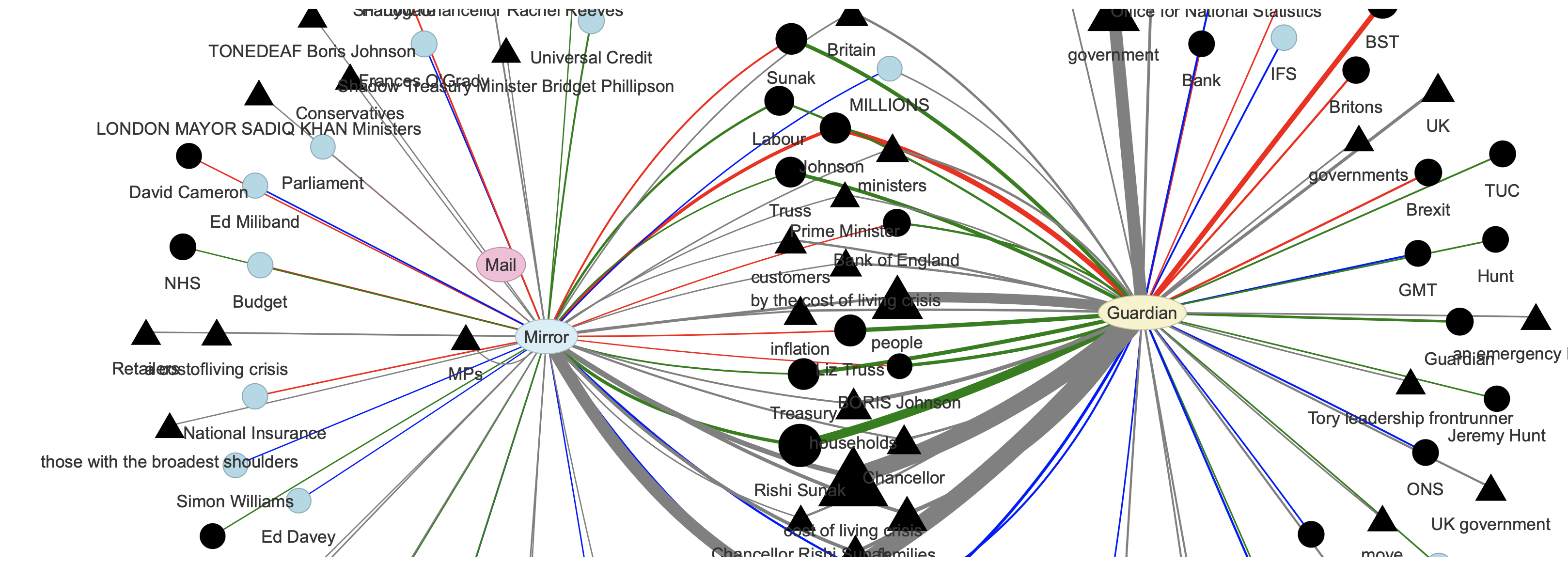}
\caption{The `cost-of-living-crisis' issue and actors associated with it.} \label{costoflivingcrisis}
\end{figure}

The pipeline picks up `cost-of-living-crisis' as a separate issue. The crisis framing seems more entrenched on the political left, since this issue links only to the Mirror and The Guardian, see Figure \ref{costoflivingcrisis}. 

Although they share the crisis framing, the Mirror and The Guardian don’t always share the sentiment associated with the different actors, e.g. here `Boris Johnson' receives positive sentiment from The Guardian, but negative one from the Mirror, and a similar pattern obtains for `the Bank of England' and `Sunak' (although it should be noted that under different naming variations, the same actors may be associated with different sentiments, e.g. `Johnson' is associated with negative sentiments for both newspapers). 
  
\begin{figure}
\includegraphics[width=\textwidth]{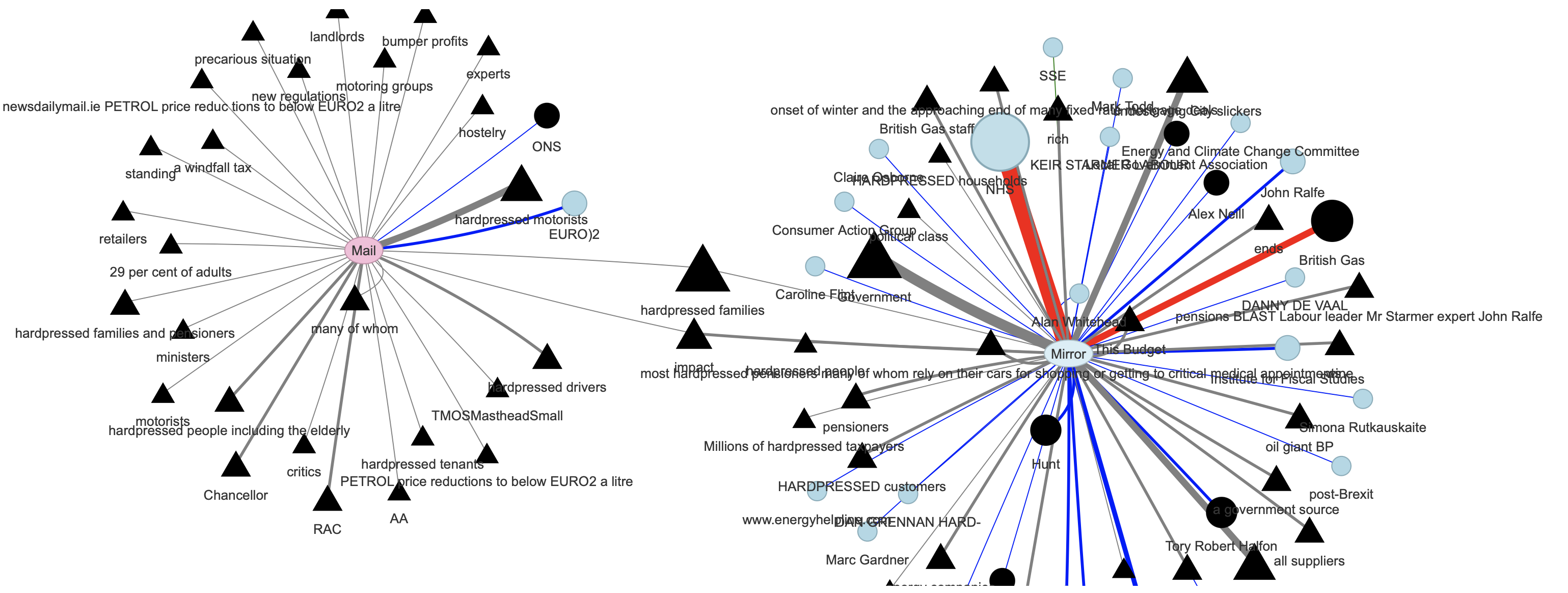}
\caption{The `hard-pressed' issue and actors associated with it.} \label{hardpressed}
\end{figure}

Other issues related to the cost of living are shared not by the political left, but by the tabloids, e.g. `hard-pressed' (shown in Figure \ref{hardpressed}), `eye-watering', `price hike'.  

\begin{figure}
\includegraphics[width=\textwidth]{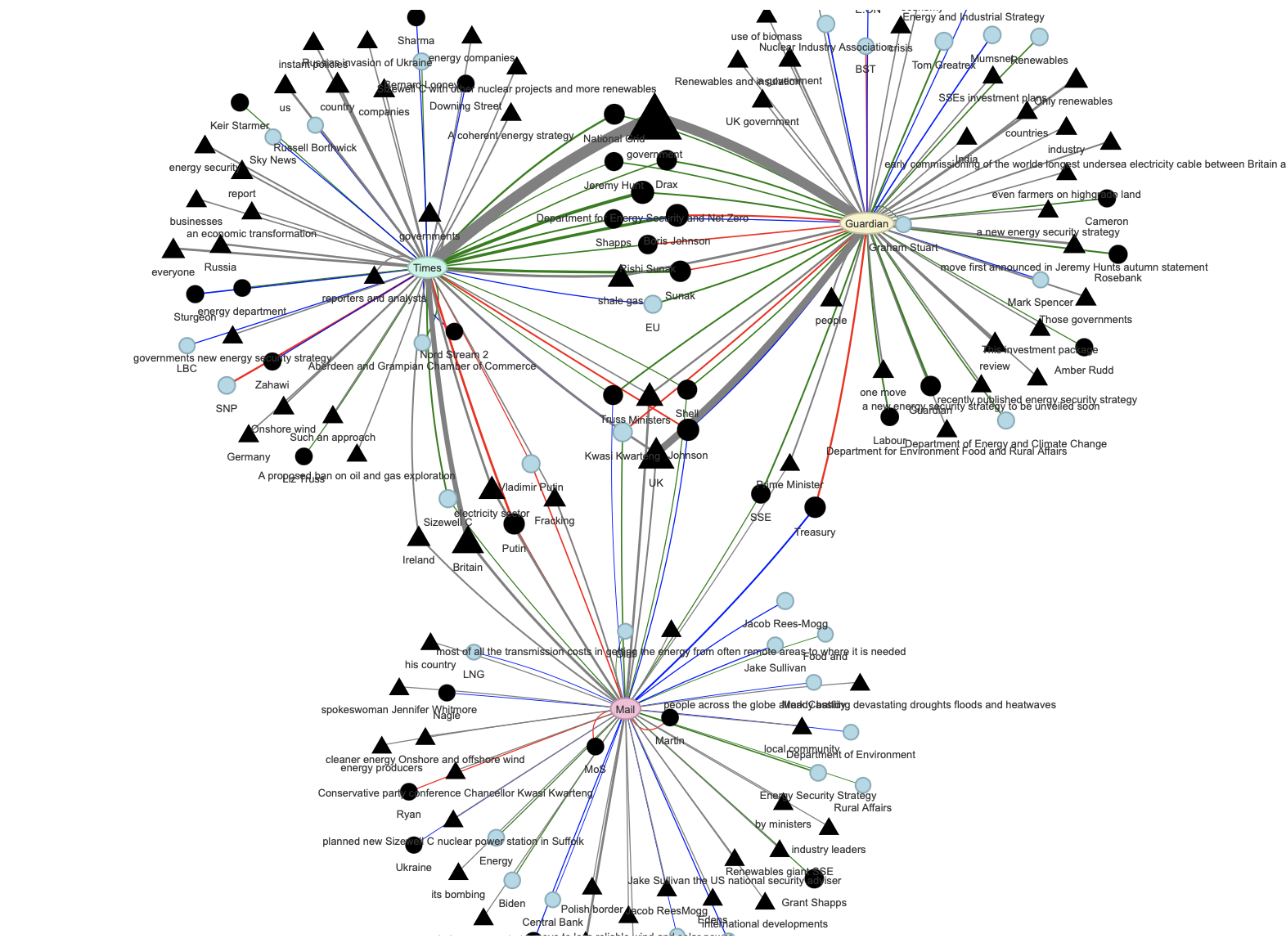}
\caption{The `energy-security' issue and actors associated with it.} \label{energysecurity}
\end{figure}

Clearly, the impact of the energy crisis and cost-of-living crisis more generally on various sections of society is given more prominence in the tabloids. The broadsheets, by contrast, where there is a wider range of topics and more attention on issues related to the economy more generally, seem to present in more detail the complex interleaving of economic and political factors. The sentiment analysis suggests that at this point negative sentiments directed at the government of the day are more readily expressed further to the left of the political spectrum, in the Mirror. Another important aspect of the energy crisis, that of energy security, is picked up as an issue too, but the Mirror is again different from the other newspapers, in that it gives it much less attention, see Figure \ref{energysecurity}. 

\begin{figure}
\includegraphics[width=\textwidth]{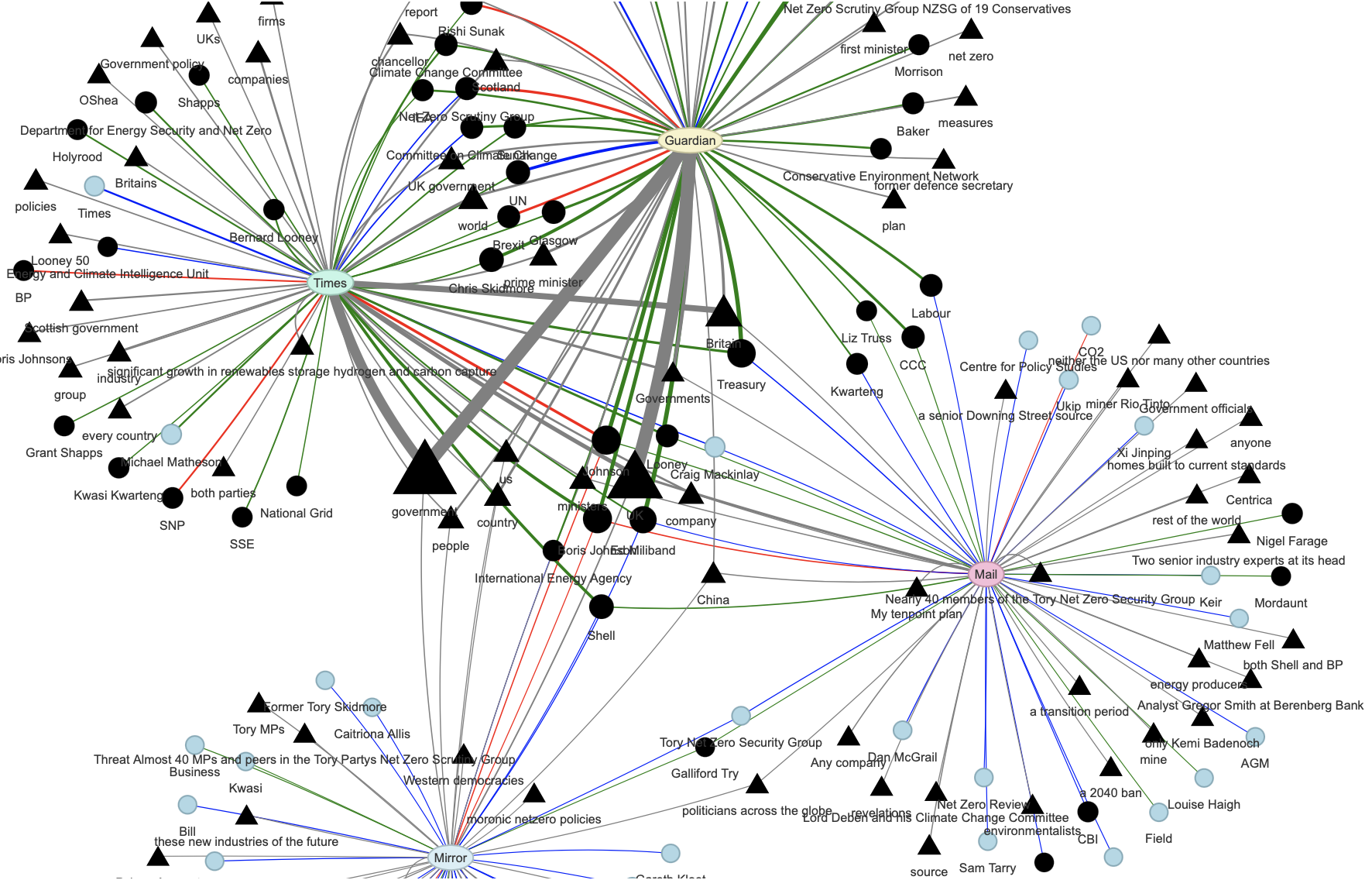}
\caption{The `net-zero' issue and actors associated with it.} \label{netzero}
\end{figure}

As noted, the analysis also picks up some issues related to the climate crisis, although this is a separate topic only in The Guardian. The range of issues we find suggests that the energy crisis gives an impetus to discussions of replacements for fossil fuels, that is, green energy and renewables. One popular issue shared across all newspapers is net-zero, see Figure \ref{netzero}. 

As before, the discussion seems to be focused on the political actors that shape policies, i.e. actors like the government, Boris Johnson, Ed Miliband, and others. The sentiment analysis suggests some expected political divisions (e.g. the Net Zero Scrutiny Group attracts negative sentiment from The Guardian and neutral sentiment from The Times) and some less expected ones (Johnson is associated with positive sentiment in the Mail but also The Guardian, and negative from the other two newspapers). This indicates that there is no complete political consensus when it comes to net zero targets, something that recent political events seem to confirm. 

\section{Discussion} \label{discussion}

In this work, we have adopted multiple frames of reference: applied at the entity level, focusing on actors and roles, and at the document level, focusing on topics and issues. Reliably identifying issues has proven to be a deceptively complex problem: the term ‘issue’ may have an intuitively simple meaning and denotation, to the extent that few human readers would have trouble recognising their existence within a sample of online debate. However, operationalising that concept into a definition that can be accurately and unambiguously applied in an automated manner is a very different undertaking. Issues can be articulated explicitly or referred to only indirectly, and often display a fractal nature within a complex inter-connected network of concepts and ideas, relying on the reader’s knowledge of world events and current affairs for their accurate interpretation. In practice, we have relied on certain heuristics to signal their presence, such as instances of common phrases that are simultaneously associated with both a high degree of positive and negative sentiment (suggesting that they have the effect of polarising the debate). This has proven to be a useful first approximation, and no doubt more robust techniques could be the focus of future work. 

As we observed above, the NLP analysis gives us an overview of what is discussed in the media. 
We can see from the topic modelling that during the energy crisis, the media was focused mostly on internal politics and internal and relevant external economic issues. The newspapers we studied framed the energy crisis primarily as a cost-of-living crisis, in some cases almost exclusively so. Climate change appears mostly as a solution to the problem, i.e. green and renewable energy, though policy discussions around net zero, for instance, appear in all newspapers. In this case, the discussion again centers around political actors and differing sentiment suggests consensus has not been reached. The presence of the climate crisis as a topic in The Guardian indicates it continued to give this significant attention. Of course, we should acknowledge that, inevitably, the selection of a particular keyword set will prioritise a particular perspective on a topic, possibly at the expense of others (such as sustainability, alternative energy sources, etc.). We recognise that multiple perspectives exist and these could be explored further as part of a future iteration.

Although the NLP tools we deployed give us a breadth of analysis, their focus remains on what is being said and who is being talked about. In the current pipeline, they don’t tell us enough about the how and the why of discourse. In some cases, the analysis retains too high a level of generality (topics), and in others, it is too granular. Semantic roles, for instance, tell us whether entities are actors or themes or both on the sentence level, but they can’t answer questions like ‘Who is held responsible for the crisis? Who is tasked with finding solutions for this crisis?’. For questions like these, and for a more fine-grained analysis of the discourse itself, we need to go to the next step, i.e. finding principled ways to extract data for qualitative analysis. A preliminary qualitative investigation of discussions around the net zero policy, for instance, suggests that The Guardian coverage during the period of the energy crisis reflects on the difficult trade-offs between providing support to people during a cost-of-living crisis, securing sufficient energy to cover energy needs in the short and the longer term, and energy security. The Guardian seems to lobby that the moment be seen as an opportunity, as encapsulated in the following quote from April 2022 suggests: “Given a cost-of-living crunch caused by the rocketing price of fossil fuels, and the new priority of energy independence following Russia's invasion of Ukraine, an imaginative and proactive government would move to [...] seize the moment.” By contrast, The Times seems to discuss net zero mostly in the context of finance and business, observing economic costs and benefits, but with a less pro-active stance with respect to climate change action. 

This brings us to the issue of evaluation. The qualitative, human-led analysis could provide a useful check on the view of the energy discourse generated by the NLP pipeline. For instance, as we mentioned above, humans have powerful intuitions about what ‘issues’ are present in media discussions. However, humans are biased and the amount of text they can read is significantly smaller. Our next step, then, would require us to integrate the manual qualitative analysis into the pipeline in a principled way that would provide for cross-fertilisation and mutual validation of the manual and automatic analyses. Part of this could include exploration other frames of reference, not just in terms of topical scope but also granularity: from broad perspectives to narrow issues and vice versa, with further investigation of the iterative retrieval approaches described in Section 3.2.

\section{Summary and conclusions}

In our work on the project to date we have built a functioning pipeline for the NLP analysis of media discourse, focusing our attention on a prominent recent media discussion: the energy crisis that affected the UK and most of the rest of the world and became particularly acute with the war in Ukraine. The pipeline gives us broad insights into the data, building a picture of the topics covered by different media sources, the issues those topics comprise, as well as the entities involved, overlaid with sentiments expressed towards them. The picture that emerges is a focus on the energy crisis as a national, rather than global, crisis, with a focus on its ramifications on people and households. Political actors are the ones most prominent in the discussion, presumably because they are tasked with resolving it. Where climate issues feature in the debate the focus is not on energy consumption as one of the causes of climate change, but on solutions and mitigations, e.g., renewable energy. 
In future work, we hope to apply these techniques to an additional corpus of social media data extracted from Twitter (now X) and compare the insights generated to highlight key points of differentiation with mainstream media. We hope also to develop partnerships with related organizations, such as the British Ecological Society, Friends of the Earth, National Association for Environmental Organisation. Finally, we hope to offer a reusable platform to support other discourse analysis investigations, e.g., public discourses around energy, poverty and equality, attitudes to risk, and more. 

\backmatter

\bmhead{Acknowledgments}

We gratefully acknowledge the financial support from the Strategic Research Fund of Goldsmiths, University of London.


\bibliography{bibliography}

\end{document}